\documentclass{article}

\usepackage[preprint]{neurips_2021}
\usepackage[english]{babel}
\usepackage[utf8]{inputenc} 
\usepackage[T1]{fontenc}    
\usepackage{hyperref}       
\usepackage{url}            
\usepackage{booktabs}       
\usepackage{amsfonts}       
\usepackage{nicefrac}       
\usepackage{microtype}      
\usepackage{xcolor}         
\usepackage{amsmath, amssymb}
\usepackage[ruled, vlined, linesnumbered ]{algorithm2e}
\usepackage{graphicx}
\usepackage{subcaption}
\DeclareMathOperator{\E}{\mathbb{E}}
\DeclareMathOperator*{\argmax}{arg\,max}

\hypersetup{
    colorlinks=true,
    linkcolor=blue,
    filecolor=magenta,      
    urlcolor=blue,
    pdftitle={Overleaf Example},
    pdfpagemode=FullScreen,
    citecolor=black
    }

\title{An Exploration of Deep Reinforcement Learning Methods with Hungry Geese}

\author{
  Matthew Kluska* \\
  Boston University\\
  \texttt{kluska@bu.edu} \\
   \And
   Nikzad Khani* \\
   Boston University \\
   \texttt{nikzad@bu.edu} \\
}

\begin{document}

\maketitle
\def\thefootnote{*}\footnotetext{These authors contributed equally to this work}\def\thefootnote{\arabic{footnote}}

\begin{abstract}
  Hungry Geese is a n-player variation of the popular game snake. This paper looks at state of the art Deep Reinforcement Learning Value Methods. The goal of the paper is to aggregate research of value based methods and apply it as an exercise to other environments. A vanilla Deep Q Network, a Double Q-network and a Dueling Q-Network were all examined and tested with the Hungry Geese environment. The best performing model was the vanilla Deep Q Network due to its simple state representation and 
  smaller network structure. Converging towards an optimal policy was found to be 
  difficult due to random geese initialization and food generation. Therefore 
  we show that Deep Q Networks may not be the appropriate model for such a stochastic
  environment and lastly we present improvements that can be made along with more 
  suitable models for the environment.
\end{abstract}

\section{Introduction}

\quad Reinforcement Learning, a branch in machine learning, deals with training some agent or a model to learn an optimal policy or function. With the introduction of open source game environments such as the Arcade Learning Environment presented by \citet{Bellemare_2013} or OpenAI's gym environments presented by \citet{brockman2016openai}, it has become the status quo to test general reinforcement learning algorithms in the baselines set force by the community. The goal of this paper is to illustrate attempts to apply general reinforcement learning algorithms to a new environment set forth by Kaggle in their Hungry Geese competition\footnote{https://www.kaggle.com/c/hungry-geese}. Since the hungry geese environment is discrete as described in section \ref{sec:state_rep} it is an optimal candidate to use Q-learning as described by \citet{Watkins1989LearningFD}. However, since the introduction of neural networks showing an increased boost in performance. Furthermore, storing values in a Q-table would be computationally inefficient as the state space is sufficiently large. One would have to store ${77 \choose 4} \approx 5.4$ million Q-values for taking values from the initial state space alone in a game starting with $4$ geese. Thus, we opted to start with a Deep Q-network as proposed by \citet{mnih} from Deepmind in section \ref{sec:dqn}. Then we added improvements mainly conversion to a Double Q-network proposed by \citet{vanhasselt2015deep} and Dueling Q-network as stated by \citet{wang} in sections \ref{sec:ddqn} and \ref{sec:dueling_dqn} respectively. 

\section{Related Work/Literature Review}

\quad Deep Reinforcement Learning was introduced by Deep Mind when they
were able to train an agent by playing Atari with the pixels on the
screen \citet{mnih2013playing} . They combined aspects from
Convolutional Neural Networks and traditional Q learning value methods
to achieve superhuman in 30+ Atari games, most notably Cartpole and
space invaders. We initially devised our literature review into 3
categories: Value Based Methods, Policy Based Methods and Model Based
Methods \citet{Fran_ois_Lavet_2018}. In the end, we kept our focus on
Value Based Methods due to the discrete nature of our environment. DQNs
were introduced and used to capture the state space thru Convelutional
Network layers and use fully connected layers to approximate a Q
function. Double DQNs like those mentioned in a Sungka Ai paper aimed at
remedying the overestimating nature of a single DQN 
\citet{bautista2019mastering}. Dueling DQNs expand further on this
concept of providing stability to the Q function by separating the
Action and State functions into two competing artificial neural networks
\citet{wang}.

\section{Methods}
\subsection{State Representation}\label{sec:state_rep}

\quad The hungry geese environment\footnote{https://github.com/Kaggle/kaggle-environments} from Kaggle will give the current positions of all the geese, and food as indices into into a 7 by 11 grid. These indices are within the range from $0$ to $76$, and we can convert these indices into a row index by integer division by the number of columns (in this case 11), and taking indices modulus the number of columns (in this case 11) to get the column index. The competition pits four geese against each other at a time including the player so we construct one-hot encodings for each position in the 7 by 11 grid as shown by Table \ref{table:encoding}.

\begin{table}[h!]
\centering
\begin{tabular}{ |c|l| } 
 \hline
 Index & Encoded item\\
 \hline
 0 & Player Head Position \\
 1-3 & Enemy Head Position \\
 4 & Player Tail Position \\
 5-7 & Enemy Tail Position \\
 8 & Player Entire Body Position \\
 9-11 & Enemy Entire Body Position \\
 12 & Player Previous Head Position \\
 13-15 & Enemy Previous Head Position \\
 16 & Food Position\\
 \hline
\end{tabular}
\caption{One-hot encodings for grid position}
\label{table:encoding}
\end{table}

The indices in Table \ref{table:encoding} represent the index into the channel dimension
of the input, where the input will be of size (Batch Size, 17, 7, 11). The DQN used a slightly simplified schema of the above encodings. Instead, the DQN treated the entire goose as one encoding, and all enemy geese as one encoding. Thus, resulting in only three one hot encodings with a input size of (Batch Size, 3, 7, 11).

\subsection{Rewards}
\quad The Kaggle environment will return a reward after an episode. This reward is calculated according to the vanilla reward algorithm shown in algorithm 1. The reward for a specific step in the episode is the difference between the current
cumulative reward and the previous steps cumulative reward. In other words the reward
for an action is given by the length of the goose added with the episode number in the state following the action.
\begin{algorithm}[H]
\label{algorithm:vanilla}
\caption{Vanilla Reward}
cumulativeReward = 0 \\
\For{$i$ \text{in episodes} }{
    reward = $length(Goose) + i$\\
    cumulativeReward += reward\\
}
\KwRet{cumulativeReward} 
\end{algorithm}
Algorithm 2 gives the reward after an episode which was used
in the training for the DQN model's policy update. 
\pagebreak
\begin{algorithm}[h!]
\caption{DQN Training Reward}
CumulativeReward=0 \\
\For{$i$ \text{in episodes}}{
        \If{(CumulativeReward == 1) and (reward != 0)}{ 
            $CumulativeReward$+= 50\\
        } 
        \If{(CumulativeReward == 0) or (len(geese[0])==0)}{ 
            $CumulativeReward$-=1000\\
            }
        \If {(max([len(goose) for goose in geese[1:]]) == 0) and (reward != 0)}{ 
            $CumulativeReward$+=1000\\
            } 
        \If {(CumulativeReward module 100)==0}{
            $CumulativeReward$+= 50\\
            } 
      \Else{ 
            $CumulativeReward$+= 10\\
            }
}
\textbf{return} $CumulativeReward$\\
\end{algorithm}

Algorithm 3 was thought to lead to dramatic improvements due to applying a direction and magnitude to the reward; however, there were little to
no changes to the double network architectures. The vanilla deep q-network was not tested for this reward function and we hope to evaluate it in the future. The 
idea with this algorithm is the closer the goose gets to the food the greater the
reward quadratically. However, going in the opposite direction to the food will lead to a linearly increasing negative reward depending on the distance from the food. The idea with this is that the risk of pursuing a closer food outweighs any danger of collision;however, when the food is far away the goose should focus on survival as
another goose will likely consume the food by the time the player goose reaches the food.

\begin{algorithm}
\label{algorithm:manhattan}
\caption{Manhattan-Based Reward}
Goose: Goose position at time step $i$\\
Goose': Goose position at time step $i - 1$\\
Food: List of food positions at time step $i$\\
Food': List of food positions at time step $i - 1$\\
MaxFoodDistance: The farthest distance the food could be from the goose\\
\BlankLine
cumulativeReward = 0\\
\For{step in episode}{
    \If{length(Goose) == 0}{
        reward = -1000
    }
    \If{$i \neq 1$ and length(Goose) > length(Goose')}{
        reward += 500
    }
    lastDistancetoFood = $\min_{f' \in \text{Food'}}{manhattanDistance(\text{Goose}', f')}$\\
    currentDistancetoFood = $\min_{f \in \text{Food}}{manhattanDistance(\text{Goose}, f)}$\\
    \BlankLine
    \eIf{\text{currentDistancetoFood  > lastDistancetoFood}} 
    {
        reward += $(\text{MaxFoodDistance} - \text{currentDistancetoFood})^2$
    }
    {
        reward -= $\text{MaxFoodDistance} - \text{currentDistancetoFood})$
    }
    cumulativeReward += reward
}
\KwRet{cumulativeReward} 
\end{algorithm}

\subsection{Deep Q-Network}\label{sec:dqn}
\quad Deep Q-networks are model free so they do not rely on a specific network architecture. 
The network architecture we used to learn the parameter $\theta_i$ consisted of  2  convolutional layers and  2  fully connected layers

\begin{equation}\label{sec:dqn_loss}
    L_i(\theta_i) = \E_{s, a, r, s'}\sim U(D)[(r + \gamma \max_{a'}{Q(s', a'; \theta_i}) - Q(s, a; \theta_i))^2]
\end{equation}

Equation \ref{sec:dqn_loss} shows how the mean-squared loss is calculated given $\theta_i$.
$s$, $a$, and $r$ are the state, action and reward of the action for the current time step, while $s'$ and $a'$ are the state and action of the following time step.

\begin{figure}[h!]
    \centering
    \includegraphics[width=0.75\textwidth]{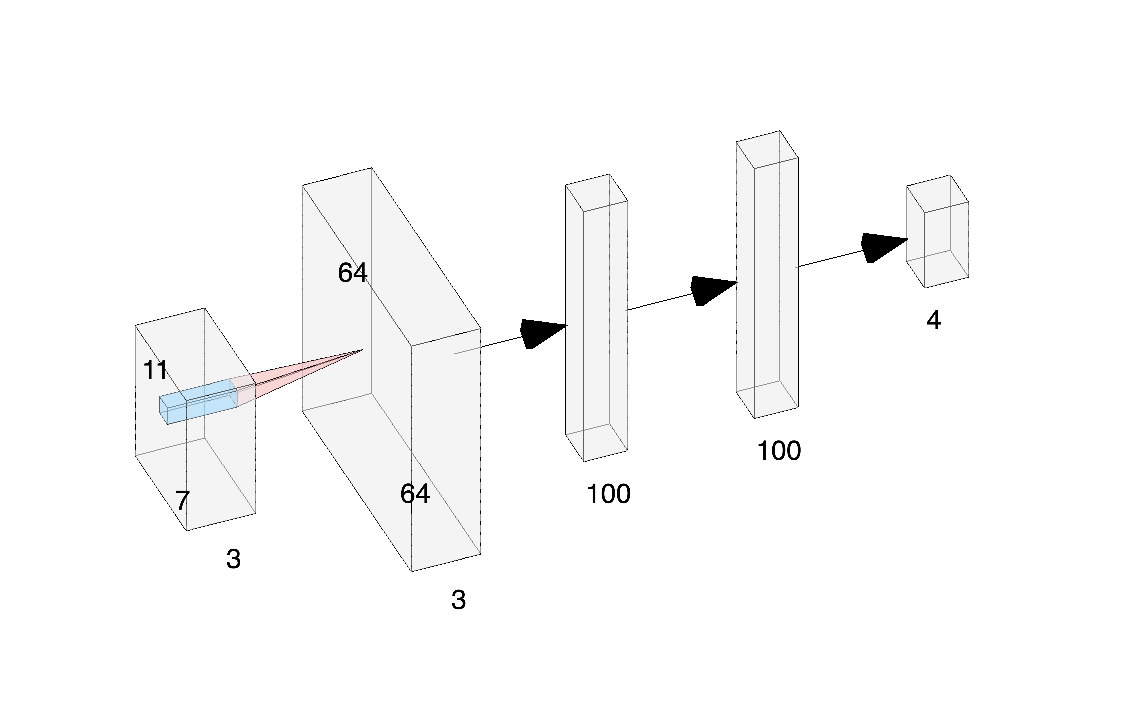}
    \caption{DQN architecture}
    \label{fig:dqn}
\end{figure}

\subsection{Double Q-Network}\label{sec:ddqn}
\quad \citet{vanhasselt2015deep} explains that Deep Q-networks have a inherent tendency to
overestimate Q-values due to the max operator in equation \ref{1}. They propose that
adding another set of weights can help reduce the overoptimism. The evaluation network will only update its weights, $\theta_k'$ every $Q_iter$ training steps. This is a hyper parameter that can be tuned for the specific environment/model. The target network's weights, $\theta_k$ on the other hand will update every training step similar to Deep Q-Networks.

\begin{align}
    q &= Q(s, a; \theta_k)\\
    \hat{q} &= R_{t+1} + \gamma Q(s', \argmax_{a \in A}Q(s, a; \theta_k); \theta_k') \\
    L_\delta(q - \hat{q}) &= \begin{cases}
                            \frac{1}{2}(q - \hat{q})^2 & |q_t - \hat{q}_t| \leq \delta\\
                            \delta(|q - \hat{q}| - \frac{1}{2}\delta)^2 & \text{otherwise} 
                        \end{cases}\label{eq:huber}
\end{align}

The network for the Double Q-Network consisted of three convolutional layers with channels (input-output) of 17-128, 128-256, 256-128. The first layer had a kernel size of (3,5) while the following two had a kernel size of (3,3). All convolutional layers
had a kernel size of were followed by batch normalization layers, and a leaky ReLU activation. The output from the last convolutional layer was flattened everywhere but the batch size was fed into two fully connected layers of size 384-64 and 64-4. The fully connected layers also had a leaky ReLU activation. The loss was evaluated with the Smooth L1 also known as the Huber Loss as shown in equation \ref{eq:huber} as proposed by \citet{huber} for its greater robustness which helps mitigate exploding gradients.

\begin{figure}[h!]
    \centering
    \includegraphics[width=\textwidth]{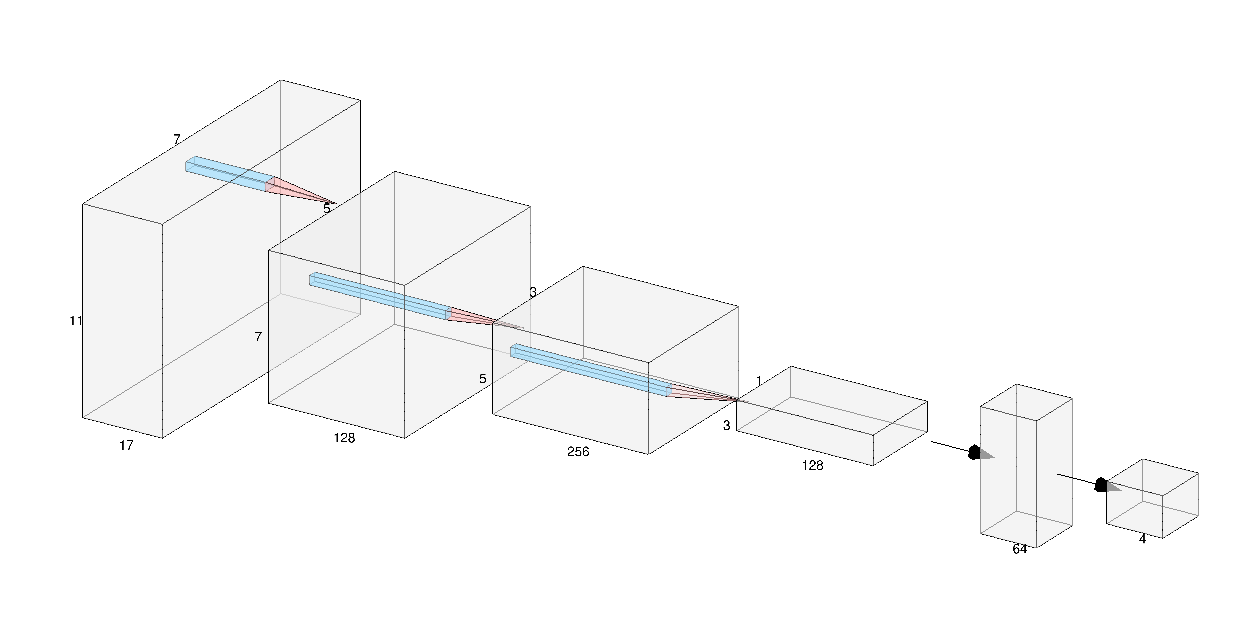}
    \caption{Deep network used for Double Deep Q-Network architecture}
    \label{fig:ddqn}
\end{figure}

\subsection{Dueling Q-Network}\label{sec:dueling_dqn}
\quad Dueling Q-Networks use the same Q-update policy as Double Q-networks ; however, \citet{wang} proposes a different network architecture. In their architecture they use convolutional layers to get an encoding for 
the state representation. The output of the convolutional layers is then inputted to two different streams of fully connected layers. The first stream
is used to learn the advantage function, $A(s,a)$ as defined by equation \ref{eq:advantage}. The advantage function will return a vector with length of the
action space. Each entry in this vector gives a value for the importance of taking
this action. The second stream is used to learn the value function as defined by equation \ref{eq:value} and will result in a single scalar. 

\begin{align}
    Q(s,a) &= \E[R_t| s_t=s, a_t = a, \pi]\\
    V(s) &= \E_{a \sim \pi(s)}[Q(s,a)]\label{eq:value}\\
    A(s,a) &= Q(s,a) - V(s) \label{eq:advantage}
\end{align}

The two streams are then combined together to get a vector of Q-values with the
length of the action space. \citet{wang} proposes that the combination
strategy in equation \ref{eq:q_combination} shows the best performance, and additionally
states that since we are only using estimates of the advantage and value functions
we cannot add them to get our Q-values, instead we require a different aggregation
strategy that will ensure expectation of the advantage is 0.

\begin{equation}\label{eq:q_combination}
    Q(s,a; \theta, \alpha, \beta) = V(s; \theta, \beta) + (A(s,a,\theta,\alpha) - \frac{1}{|\Lambda|}\sum_{a' \in \Lambda}A(s, a';\theta, \alpha))
\end{equation}

\quad The Q function in equation \ref{eq:q_combination} is parametrized by the weights of the convolutional layers, $\theta$, the weights of the advantage function stream, $\alpha$, and the weights of the value function stream, $\beta$. $\Lambda$ is the action
space.

    \quad The neural network used for dueling consisted of the same convolutional
and batch normalization layers as the Double Q network. The output of the last
convolutional layer was instead fed into two different 384-128 fully connected
layers with leaky ReLU activation. The first fully connected layer was followed by 
another 128-1 fully connected layer for the value function, and the second fully connected layer was followed by a 128-4 fully connected layer for the advantage function. The output of the 128-1 and 128-4 were aggregated using equation \ref{eq:q_combination}, and evaluated with Huber Loss as well shown in equation 
\ref{eq:huber}.

\section{Results}

\quad The DQN showed continuous improvement over the course of the experiment. It was trained for 24 hours on a single GPU instance and we saw a steady decrease in loss, increase in score and increase in elo. After this 24 hours of training and roughly 50k iterations, it was able to achieve a win rate of 0.2 and a max cumulative reward of 300. We believe that there is room for our DQN to achieve higher results if it is given more iterations( roughly 5-10M) to train. This way it can continuously update its Q network and descent towards the optimal policy.

\quad The Dueling network architecture was unable to learn/converge towards an optimal
policy due to reasons described in section \ref{sec:discussion}. Its initial good performance was due to choosing a rule-based agents actions during exploration in
the epsilon-greedy training cycle. This was implemented in an effort to have the
Q-network memorize the greedy actions then improve them through random exploration
later in training; however, over 1000 iterations the model was not able to effectively
learn the policy leading to a severe drop in performance as it begins exploring with
random actions and greedily choosing actions from the network.

\begin{figure}[h!]
\begin{subfigure}{0.5\textwidth}
    \centering
    \includegraphics[width=0.9\linewidth]{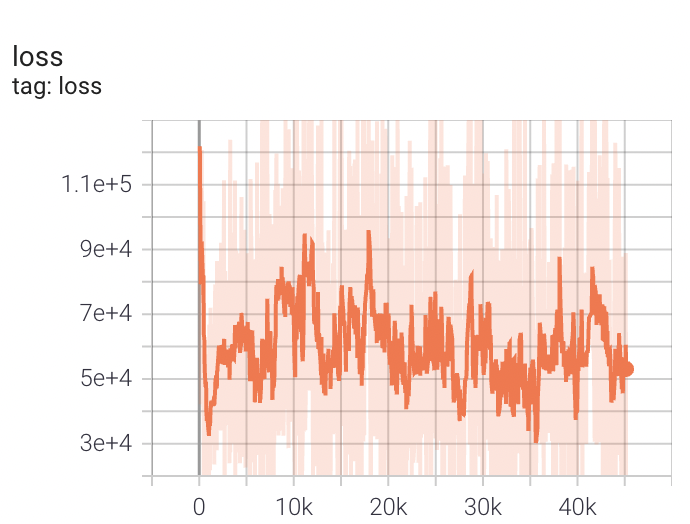}
    \caption{DQN MSE Loss}
    \label{fig:dqn_mse_loss}
\end{subfigure}
\begin{subfigure}{0.5\textwidth}
    \centering
    \includegraphics[width=0.9\linewidth]{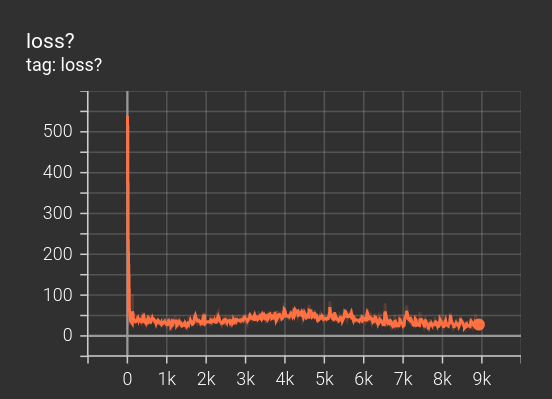}
    \caption{DDQN Huber Loss}
    \label{fig:ddqn_huber_loss}
\end{subfigure}
\begin{subfigure}{0.5\textwidth}
    \centering
    \includegraphics[width=0.9\linewidth]{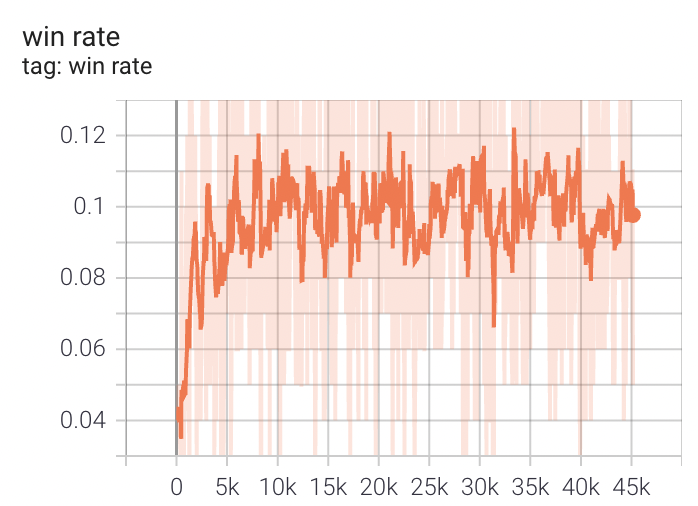}
    \caption{DQN Win Rate}
    \label{fig:dqn_win_rate}
\end{subfigure}
\begin{subfigure}{0.5\textwidth}
    \centering
    \includegraphics[width=0.9\linewidth]{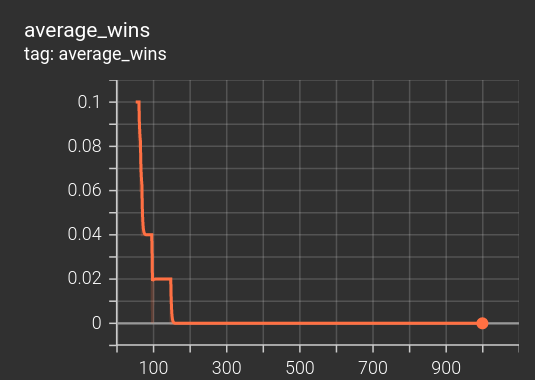}
    \caption{DDQN Win Rate}
    \label{fig:ddqn_win_rate}
\end{subfigure}
\begin{subfigure}{0.5\textwidth}
    \centering
    \includegraphics[width=0.9\linewidth]{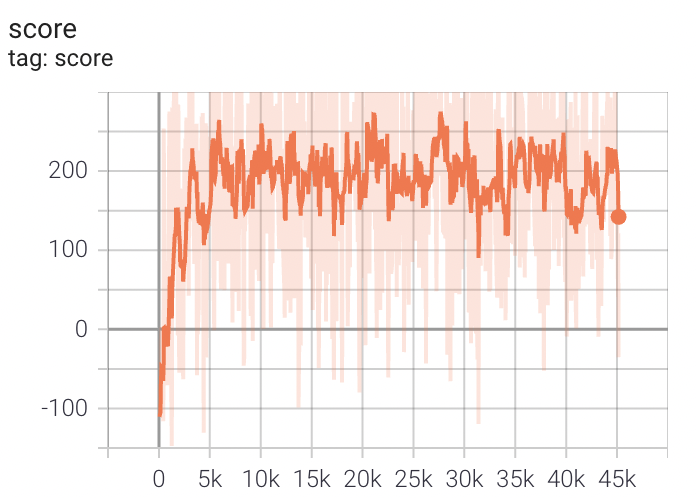}
    \caption{DQN Score}
    \label{fig:DQN_score}
\end{subfigure}
\begin{subfigure}{0.5\textwidth}
\centering
    \includegraphics[width=0.9\linewidth]{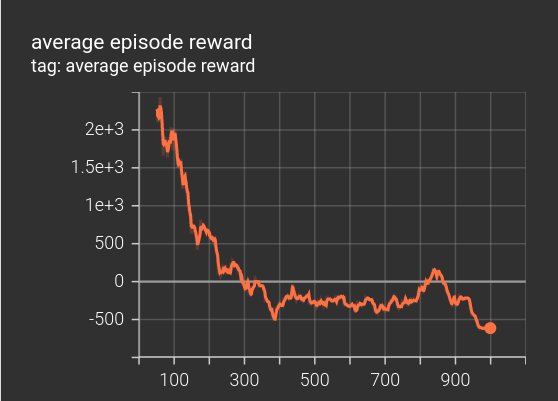}
    \caption{DDQN Score}
    \label{fig:DDQN Score}
\end{subfigure}

\end{figure}

\section{Discussion}\label{sec:discussion}
\quad Q networks are able to converge fastest in an environment that is completely deterministic. An example of this would be trying to learn how to navigate 
through a maze. When you are able to reach the same state simply through greedy actions
of the learned policy then exploration up to that point becomes unnecessary to succeed.
However, a much trickier problem is to learn how to navigate in a maze where we are
randomly dropped somewhere in the maze. This will require a much more thorough understanding of the state space, despite the state space remaining the same size 
as the previous example. This is because we are coerced to learn paths from
states that we would previously not visit. Furthermore, learning the optimal path 
from a random state would require that we visit that state again to learn the best next action. An even more difficult problem to learn the optimal path is for when the end
of the maze is also randomized every episode. Because this can render the previously
learned paths completely useless. This situation is analogous to the hungry geese environment. 

\quad The random initialization of the geese and the food leads to a huge amount of initial state spaces and learned paths which are not even applicable for future episodes since there will be another random state again after the episode ends. We thought increasing the input size and giving the network more information will help it learn similarities between states better such as that it should not take the opposite action; however, we believe that Q-values do not converge sufficiently for this to take place. Thus, the lower win rate and performance of the dueling networks is attributed to not only learning on a larger input size but also learning on a far larger network.
\quad 

\section{Conclusion}
DQNs are not built to deal with inherently stochastic environments.  Policy gradient methods(PPO) or model based methods like monte-carlo search tree (used in Alpha Go/ Alpha Zero) would have probably been the better choice. We were still able to win against a 4 greedy agents roughly 10-20 percent of the time.
Due to the dueling architectures of DDQNs, they provided inherently more stability. Actor-Critic Models also share some similarities(Dual Neural Nets) and would have been a competing model to pursue if time allowed.

\section{Future Work}
We simplified our assumptions and focused narrowly on value-based methods deeply. This opens up room for improvement with the use of either different methods:Policy Gradient Methods( PPO, Actor Critic, etc.) and Model based Methods (Monte Carlo Tree Search algorithms) as well as innovations to augmenting the state space. 
Two theories to augmenting the state space were to: \\
\\
1. Center the goose at (6,4) \\
2. Either 90,180,270 board rotations so that the snake's last move was up and it has only 3 options.( Up, left and right)

The first innovation would take advantage of board wrapping.
The second innovation would reduce our action space and eliminate any chance of the goose killing itself in the game.

In terms of architecture improvements taking advantage of parallelization of jobs and multiple GPUs thru training architectures like IMPALA would have allowed for higher GPU utilization rates and efficiently using multiple GPUs \citet{espeholt2018impala}.Facebook research implements of version of this called torchbeast and it would allow for higher training thru-put on available hardware.

\section{Acknowledgements}
All the code for this paper can be found \href{https://www.github.com/nikzadkhani/Famished-Geese}{here}. We would like to thank
Professor Sang Chin from Boston University and Andrew Wood from Boston University for their support and help in this paper.

\newpage
\bibliography{anthology}

\begin{thebibliography}{10}
\providecommand{\natexlab}[1]{#1}
\providecommand{\url}[1]{\texttt{#1}}
\expandafter\ifx\csname urlstyle\endcsname\relax
  \providecommand{\doi}[1]{doi: #1}\else
  \providecommand{\doi}{doi: \begingroup \urlstyle{rm}\Url}\fi

\bibitem[Bellemare et~al.(2013)Bellemare, Naddaf, Veness, and
  Bowling]{Bellemare_2013}
M.~G. Bellemare, Y.~Naddaf, J.~Veness, and M.~Bowling.
\newblock The arcade learning environment: An evaluation platform for general
  agents.
\newblock \emph{Journal of Artificial Intelligence Research}, 47:\penalty0
  253–279, Jun 2013.
\newblock ISSN 1076-9757.
\newblock \doi{10.1613/jair.3912}.
\newblock URL \url{http://dx.doi.org/10.1613/jair.3912}.

\bibitem[Brockman et~al.(2016)Brockman, Cheung, Pettersson, Schneider,
  Schulman, Tang, and Zaremba]{brockman2016openai}
Greg Brockman, Vicki Cheung, Ludwig Pettersson, Jonas Schneider, John Schulman,
  Jie Tang, and Wojciech Zaremba.
\newblock Openai gym, 2016.

\bibitem[Espeholt et~al.(2018)Espeholt, Soyer, Munos, Simonyan, Mnih, Ward,
  Doron, Firoiu, Harley, Dunning, Legg, and Kavukcuoglu]{espeholt2018impala}
Lasse Espeholt, Hubert Soyer, Remi Munos, Karen Simonyan, Volodymir Mnih, Tom
  Ward, Yotam Doron, Vlad Firoiu, Tim Harley, Iain Dunning, Shane Legg, and
  Koray Kavukcuoglu.
\newblock Impala: Scalable distributed deep-rl with importance weighted
  actor-learner architectures, 2018.

\bibitem[François-Lavet et~al.(2018)François-Lavet, Henderson, Islam,
  Bellemare, and Pineau]{Fran_ois_Lavet_2018}
Vincent François-Lavet, Peter Henderson, Riashat Islam, Marc~G. Bellemare, and
  Joelle Pineau.
\newblock An introduction to deep reinforcement learning.
\newblock \emph{Foundations and Trends® in Machine Learning}, 11\penalty0
  (3-4):\penalty0 219–354, 2018.
\newblock ISSN 1935-8245.
\newblock \doi{10.1561/2200000071}.
\newblock URL \url{http://dx.doi.org/10.1561/2200000071}.

\bibitem[Huber(1964)]{huber}
Peter~J. Huber.
\newblock {Robust Estimation of a Location Parameter}.
\newblock \emph{The Annals of Mathematical Statistics}, 35\penalty0
  (1):\penalty0 73 -- 101, 1964.
\newblock \doi{10.1214/aoms/1177703732}.
\newblock URL \url{https://doi.org/10.1214/aoms/1177703732}.

\bibitem[Mnih et~al.(2013)Mnih, Kavukcuoglu, Silver, Graves, Antonoglou,
  Wierstra, and Riedmiller]{mnih2013playing}
Volodymyr Mnih, Koray Kavukcuoglu, David Silver, Alex Graves, Ioannis
  Antonoglou, Daan Wierstra, and Martin Riedmiller.
\newblock Playing atari with deep reinforcement learning, 2013.

\bibitem[Mnih et~al.(2015)Mnih, Kavukcuoglu, Silver, Rusu, Veness, Bellemare,
  Graves, Riedmiller, Fidjeland, Ostrovski, and et~al.]{mnih}
Volodymyr Mnih, Koray Kavukcuoglu, David Silver, Andrei~A. Rusu, Joel Veness,
  Marc~G. Bellemare, Alex Graves, Martin Riedmiller, Andreas~K. Fidjeland,
  Georg Ostrovski, and et~al.
\newblock Human-level control through deep reinforcement learning, Feb 2015.
\newblock URL \url{https://www.nature.com/articles/nature14236}.

\bibitem[van Hasselt et~al.(2015)van Hasselt, Guez, and
  Silver]{vanhasselt2015deep}
Hado van Hasselt, Arthur Guez, and David Silver.
\newblock Deep reinforcement learning with double q-learning, 2015.

\bibitem[Wang et~al.(2016)Wang, Schaul, Hessel, Hasselt, Lanctot, and
  Freitas]{wang}
Ziyu Wang, Tom Schaul, Matteo Hessel, Hado Hasselt, Marc Lanctot, and Nando
  Freitas.
\newblock Dueling network architectures for deep reinforcement learning.
\newblock In Maria~Florina Balcan and Kilian~Q. Weinberger, editors,
  \emph{Proceedings of The 33rd International Conference on Machine Learning},
  volume~48 of \emph{Proceedings of Machine Learning Research}, pages
  1995--2003, New York, New York, USA, 20--22 Jun 2016. PMLR.
\newblock URL \url{http://proceedings.mlr.press/v48/wangf16.html}.

\bibitem[Watkins(1989)]{Watkins1989LearningFD}
Christopher Watkins.
\newblock Learning from delayed rewards.
\newblock 1989.

\end{thebibliography}



\end{document}